\documentclass[conference,compsoc]{IEEEtran}

\usepackage[hyphens,spaces,obeyspaces]{url}
\usepackage{graphicx}
\usepackage[backend=biber,style=numeric,natbib=true,sorting=none]{biblatex}
\usepackage[inline, shortlabels]{enumitem}
\usepackage[T1]{fontenc}
\usepackage{algorithm}
\usepackage{algorithmic}
\usepackage{amsmath}
\usepackage{amssymb}
\usepackage{etoolbox,siunitx}
\robustify\bfseries
\usepackage{booktabs}
\usepackage{multirow}
\usepackage{siunitx}
\usepackage{placeins}
\usepackage{gensymb}
\usepackage[ttdefault=true]{AnonymousPro}

\IEEEoverridecommandlockouts
\IEEEpubid{\makebox[\columnwidth]{978-1-7281-3605-9/19/\$31.00~\copyright2019 IEEE \hfill} \hspace{\columnsep}\makebox[\columnwidth]{ }}

\title{\LARGE \bf
Sensor Aware Lidar Odometry
}
\author{
    \makebox[.25\linewidth]{\centering{Dmitri Kovalenko}}
    \makebox[.25\linewidth]{\centering{Mikhail Korobkin}}
    \makebox[.25\linewidth]{\centering{Andrey Minin}}\\
         Yandex LLC \\
        \centering{\tt\small \{dmk0v, m1khailk, andreyminin\}@yandex-team.ru}
}

\begin{document}
\maketitle
\IEEEpubidadjcol

\begin{abstract}
A lidar odometry method, integrating into the computation the knowledge about the physics of the sensor, is proposed.
A model of measurement error enables higher precision in estimation of the point normal covariance.
Adjacent laser beams are used in an outlier correspondence rejection scheme.
The method is ranked in the KITTI's leaderboard with 1.37\% positioning error.
3.67\% is achieved in comparison with the LOAM method on the internal dataset.
\end{abstract}

\section{Introduction} \label{sec:intro}
\begin{figure}[h!]
\centering{
\includegraphics[width=.49\linewidth]{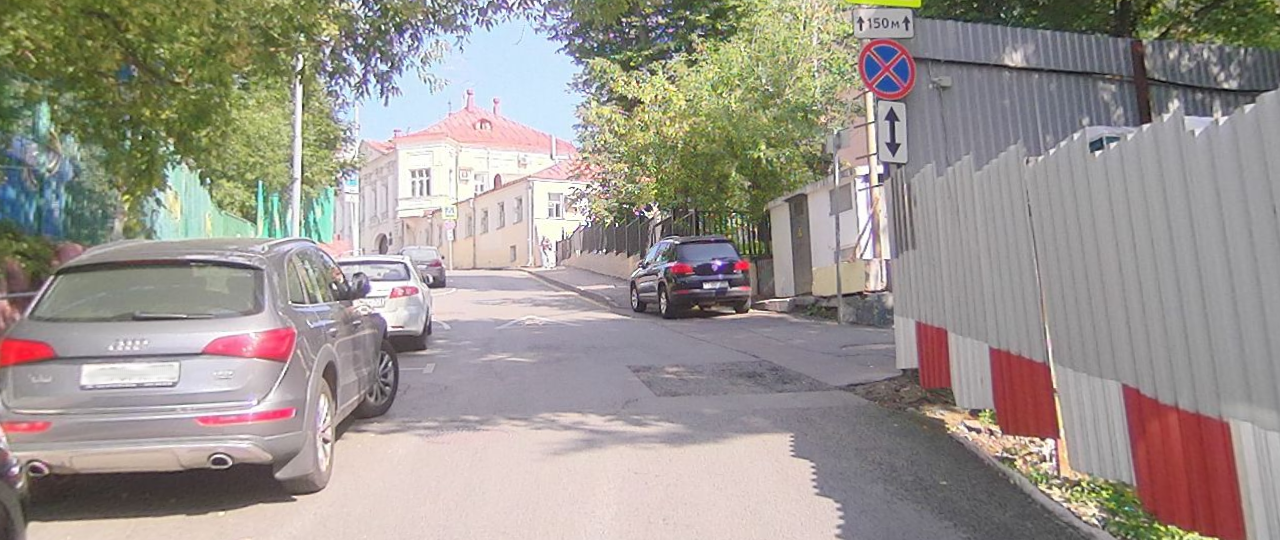}
\includegraphics[width=.49\linewidth]{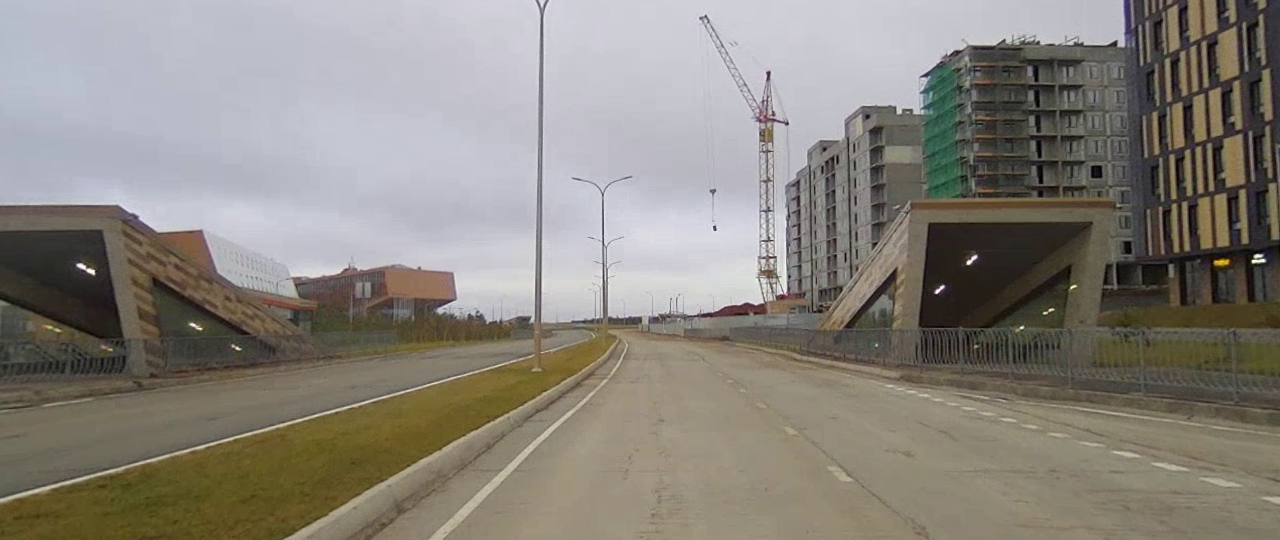}\\
\includegraphics[width=.49\linewidth]{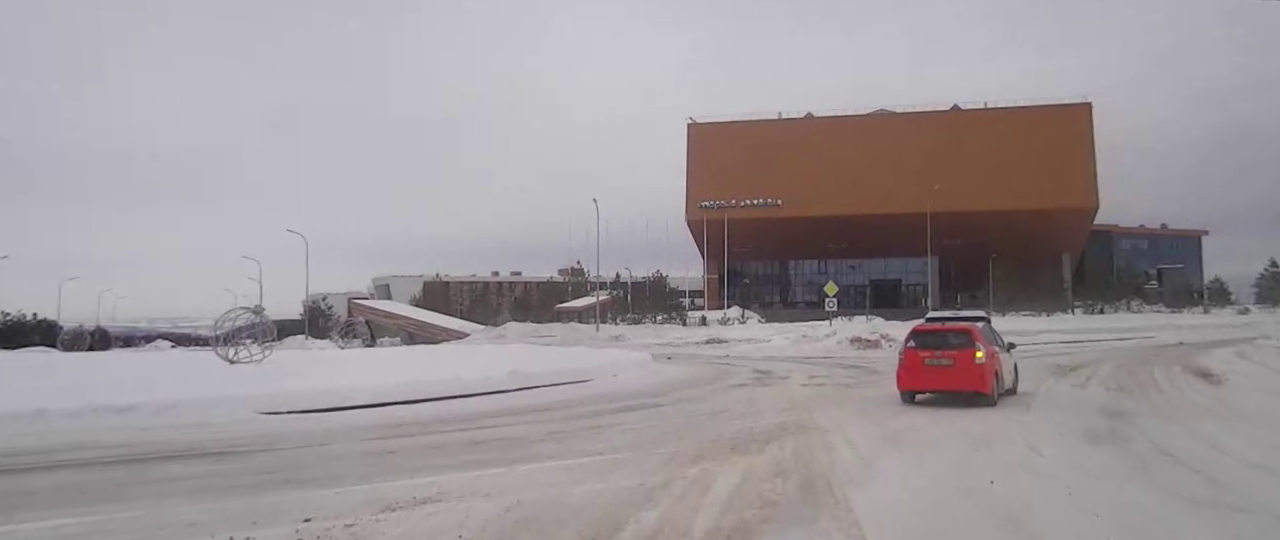}
\includegraphics[width=.49\linewidth]{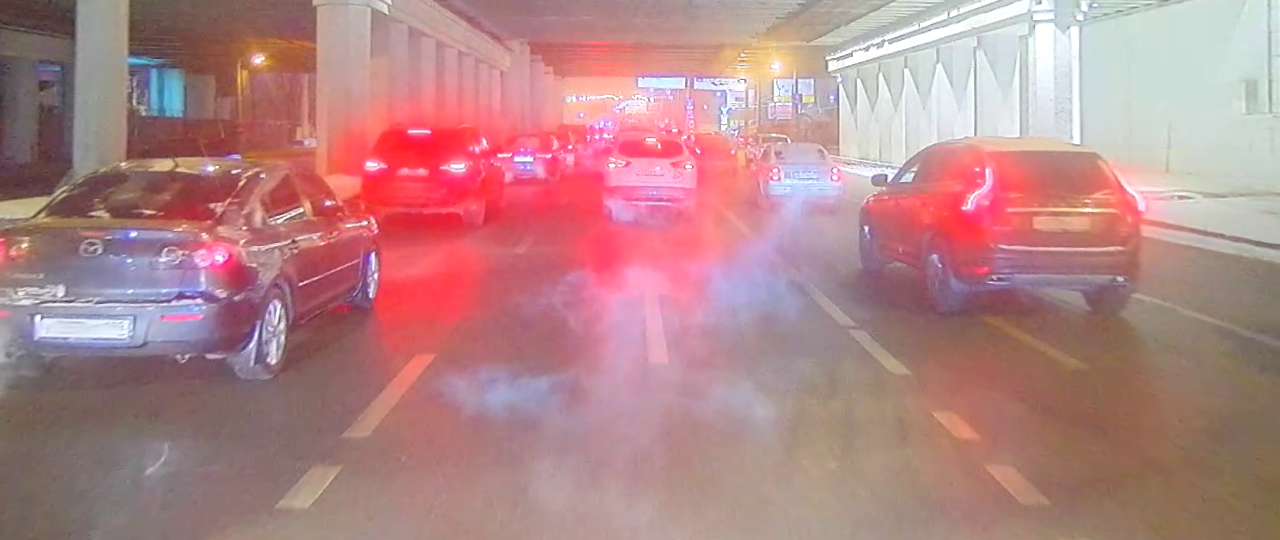}
}
\caption{Environments included in the internal dataset. Camera images are used for visualization purposes, the method operates on lidar scans only}
\label{gr:operation}
\end{figure}

The driverless transportation is becoming a technological dream on a verge of turning
  a day-to-day reality, as more competitors are joining to~\cite{yandex_in_vegas} the pursuit.
Early prototypes appeared decades ago~\cite{leonard1991simultaneous}, while DARPA
  Urban Challenge could be considered an important milestone.
AV~(Autonomous Vehicle), hardware and models of the victorious team are highlighted in~\cite{urmson2007tartan}
  with an emphasis on the crucial role that a lidar plays in providing that level of autonomy.
The comprehensive analysis~\cite{fletcher2008cornell}
  of a low-speed collision between MIT's AV "Talos" and Cornell's AV "Skynet" occured
  during the challenge puts the positioning error as a one of causes.
The lidar odometry is a subject of this work and our main contributions are:
\begin{itemize}
  \item point cloud filtering by the covariance of point normal,
    informed of the measurement error, specific for a lidar sensor
  \item scheme for the false matches rejection between consecutive point clouds,
      based on a \textit{neighbor beam distance}
  \item reliable performance in the realistic environment is demonstrated on KITTI~\cite{geiger2012we} and on
    the internal dataset, collected from the autonomous fleet
    facing intense modes of operation every day (Fig.~\ref{gr:operation}) in comparison with LOAM,
    state-of-the-art competitor~\cite{zhang2014loam}
\end{itemize}

\section{Related work} \label{sec:rel_work}
Driver-assist systems~(autonomy level $\le$ 3 \cite{SAE_ADAS}) tend
  to avoid 3D SLAM~(Simultaneous Localization and Mapping) by reliance on GNSS, odometry and the lane detection. 
However, level 4 AVs often benefit from data stored in a HD map.
This study is concerned with a level 4 setup, which dictates the separation of localization and mapping
  because doing the latter repetitively on an every car in a fleet in a same location is wasteful.

A substantial body of reseach exists on a problem of egomotion estimation and SLAM, yet
the application to AVs imposes an unique set of challenges, as the environment is:
\begin{itemize}
  \item unprepared for autonomous navigation and is unfeasible to be retrofitted with markers
  \item feature-poor: highways, tunnels, wastelands
  \item dynamic: vegetation growth, snowbanks accumulation, construction works
  \item hard to perceive: the weather, time of a day and an annual season change influence
  \item rich with occlusions of background features by other vehicles
\end{itemize}

The degenerative effect of dense urbanisation on the SLAM performance was shown in~\cite{wen2018performance}.
This challenge is addressed by graph optimization~\cite{grisetti2010tutorial}, achieving the map global consistency.
However, to reduce the computation time and decrease the chance of convergence to a suboptimal local minima, the better maps for optimization are called for. 
The input maps are built with odometry, which is the objective of our study.

The comparison~\cite{sokolov2017analysis} of lidar SLAM with the visual counterparts favours the former.
The top of the KITTI's leaderboard is also occupied by lidar methods.
Laser range measurements tend to show a higher reliability and lower dependence
  on luminance than the depth measurements produced by cameras,
  which drives our focus on a lidar.

The seminal contribution is Iterative Closest Point~\cite{besl1992method},
  a method to register a relative 3D transformation between point clouds of arbitrary shape, based on the least squares.
The closest point correspondence estimation is replaced by a grid lookup, and Euclidian loss is replaced by Mahalanobis in Normal Distributions Transform~\cite{biber2003normal}.
Novel learning-based methods are proposed~\cite{velas2018cnn},~\cite{cho2019deeplo}, but do not perform on par yet.
Reflectance intensity images with depth are reconstructed from range scans in~\cite{dong2014lighting}
  and then used as an input for the visual odometry algorithm.
Contrary to that, many works~\cite{deschaud2018imls},~\cite{neuhaus2018mc2slam},
  ~\cite{ji2018cpfg}, ~\cite{behley2018efficient}
  rely solely on the geometric properties of registered laser point clouds.

The continuous time was found benefitial by H. Alismail et al in~\cite{alismail2014continuous} as it allows
  to solve simultaneously and explicitely for the pose and the point cloud drift,
  which happens due to an egomotion during the sensor spin time.
The proposed method does not estimate the point cloud drift correction,
  as the wheel odometry data could be employed for that.

Optimal registration algorithms~\cite{cai2019practical} guarantee convergence
  without initialization, but are prohibitively expensive to compute.
The landmark-based registration is advocated in~\cite{li2010extracting} on grounds of decreasing a SLAM graph density.
In spite of this valid concern our method operates directly on point clouds,
  because the dense approach provides a greater robustness in feature-poor environments.

Experimental results are evaluated against an another feature-based approach, LOAM~\cite{zhang2014loam},
  which holds the top result on KITTI's leaderboard\footnote{http://www.cvlibs.net/datasets/kitti/eval\_odometry.php} at the time of writing this paper.
LOAM makes use of two types of features (corner-like and plane-like)
  associated with dedicated losses (point-to-line, point-to-plane).
These point-to-plane losses may include outlier entries because plane normal estimates are computed with three points each and could contain noise,
  while the proposed method estimates normals robutly and filters point clouds by normal covariances.

The rest of the paper is organized as follows:
  Sec.~\ref{sec:model} introduces the mathematical model of the proposed method,
    experimental results are layed down in Sec.~\ref{sec:experiments},
    then Sec.~\ref{sec:conclusions} concludes with the summary and further work.

\section{Lidar Odometry} \label{sec:model}
The odometry task could be stated formally as
  the problem of determining a relative rigid-body 3D transformation $T_{i}^{i-1} \in SE(3)$,
  which brings the sensor coordinate frame at a time of the current range scan (source point cloud $\widetilde{X}_{i}$) registration 
  to the sensor coordinate frame at a time of the previous range scan (target point cloud $\widetilde{X}_{i-1}$) registration.

The proposed method is built on the ICP framework~\cite{besl1992method}.
We proceed with detailing specific choices made
  for every subroutine  mentioned in Alg.~\ref{alg:icp}
  without a dive into the discussion of available alternatives,
    which are covered by the excellent review~\cite{rusinkiewicz2001efficient} and
    the best-practices for choosing among them~\cite{pomerleau2013comparing}.

\textbf{Initialization} of the transformation (initial guess) practically is
 derived  from the independent sensors available: IMU, wheel odometry.
However, this work considers the lidar-only odometry and
  uses the linear extrapolation from past states for initialization.

\textbf{Filtration of Points} is crucial as the data rate of a modern lidar would hinder the computational performance.
Depending on the lidar model and the configuration of the environment 30K-150K
  points are registered within one range scan $\widetilde{X}_i$, while only
  $5-15\%$ are passed on through the filter~(Alg.~\ref{alg:icp}, line~\ref{alg:icp:filter})
  to the model as the point cloud $X_i$.
The filtration steps are:
\begin{itemize}
  \item voxel grid filter, passing on centroids of occupied voxels
  \item normal estimation and normal covariance filtration, covered in detail in Sec.~\ref{sec:cov_filtr}
  \item local curvature filter: points with a higher curvature are discarded~(estimated as in~\cite{pcl_curvature})
\end{itemize}

\textbf{Matching of Points} is done with a k-d tree, selecting a closest point in the target cloud.
These preliminary correspondences undergo the rejection of outliers:
\begin{itemize}
  \item geometric correspondence rejector, highlighted in Sec.~\ref{sec:geom_rej}
  \item remove $20\%$ of matches having the largest distance (introduced by
        D. Chetverikov et al in~\cite{chetverikov2002trimmed})
\end{itemize}

\textbf{Transformation Estimation} is performed by the uniformly weighted
  linearized least squares optimization with the point-to-plane metric,
  which was shown to be generally superior to point-to-point by
    F. Pomerleau et al in~\cite{pomerleau2013comparing}.

\textbf{Termination Criteria} is a combination of thresholds on 
  the absolute and relative magnitude of transformation increment $\Delta$
  with the threshold on an iterations count.

\begin{algorithm}
  \begin{algorithmic}[1] 
    \REQUIRE $X_{i-1}$, $\widetilde{X}_{i}$ previous and current lidar point clouds,
             $\hat{T}_{i}^{i-1} \in SE(3)$ transformation initialization  
    \ENSURE $T_{i}^{i-1} \in SE(3)$, estimated transformation
    \STATE $X_i \Leftarrow \text{filterPoints}(\widetilde{X}_i)$ \label{alg:icp:filter}
    \STATE $X_i \Leftarrow \hat{T}_{i}^{i-1} \cdot X_i$
    \STATE $T_{i}^{i-1} \Leftarrow \hat{T}_{i}^{i-1}$
    \WHILE{not $\text{terminationCriteria}(k, \Delta, T_{i}^{i-1})$}
      \STATE $\widetilde{M} \Leftarrow \text{match}(X_{i-1}, X_{i})$ \label{alg:icp:match}
      \STATE $M \Leftarrow \text{rejectMatches}(\widetilde{M})$
      \STATE $\Delta \Leftarrow \text{estimateTransform}(X_{i-1}, X_{i}, M)$
      \STATE $T_{i}^{i-1} \Leftarrow \Delta \otimes T_{i}^{i-1}$
      \STATE $X_i \Leftarrow \Delta \cdot X_i$
    \ENDWHILE
    \RETURN $T_{i}^{i-1}$
  \end{algorithmic}
  \caption{Estimate the transformation between consecutive range scans with ICP}
  \label{alg:icp}
\end{algorithm}
With that the ICP framework is defined and
  the proposed models for point filtration and matching follow.

\subsection{Normal Covariance Filtration} \label{sec:cov_filtr}
The filter introduced in Alg.~\ref{alg:icp}, line~\ref{alg:icp:filter},
takes a raw range scan $\widetilde{X}_i$ and outputs $X_i$, where each point has an estimated normal associated with it, as well as the normal covariance:
$$
  \widetilde{X_i} = \cup_{k=0}^N{p_k}\quad X_i = \cup_{k=0}^M {(p_k, n_k, \text{Cov}[n_k])}
$$
$M \le N$, as the points having high-uncertainty normals are rejected.
The reason is miscalculated normals impair transformation estimation precision,
  as individual losses would be off.

The normal $n_k$ is estimated using SVD~(Singular Value Decomposition) in a $k$-neighborhood of the point $p_k$,
  so when a substantial fraction of points is sampled from an another surface,
  which could happen in many boundary cases, SVD would yeild a normal vector, that
    is not describing a true plane well.
The higher values of the last singular value are indicative of the lower confidence in the normal orientation.
However, such a model discounts the fact that the point coordinates are not certain
  and are subject to the measurement error.

We propose a normal covariance filter~(NCF),
  which accounts for the uncertainty in point coordinates, treating it as Gaussian random variables.
The measurement error is approximated by a sphere with the standard deviation $\xi$.

For the notation clarity, until the rest of the section the lower indices are used (as in $D_{i,j}$)
  to access a matrix entry and a single lower index to denote a matrix column $D_i$,
  counting indices from $0$.

The SVD input is an array of unbiased local points, sampled around $p_k$ from $\widetilde{X}_i$,
  i.e. the data matrix $D \in \mathbb{R}^{\{k \times 3\}}$.
The SVD output is $D = U S V$, where $U \in \mathbb{R}^{\{k \times 3\}}$,
$S \in \mathbb{R}^{\{3 \times 3\}}$, $\forall i,j: \{i \ne j; i,j < 3; S_{i,j} = 0\}$,
  $V \in \mathbb{R}^{\{3 \times 3\}}$ and $U$, $V$ are orthogonal.

When entries of $D$ are random variables, $V_2$ is as well.
The covariance of the plane normal is estimated by the propagation of uncertainty technique:
\begin{equation}
  \text{Cov}[V_2] = J^T \text{Cov}[D] J \quad  J = \frac{\partial f(D)}{\partial D}(D)
  \label{eq:cov_v2}
\end{equation}
The data matrix covariance was informally introduced above and is $\text{Cov}[D] = \xi \cdot I^{\{3 \times 3\}}$.
The function $f$ is a sequence of computations, yielding $V_2$ from $D$.
To avoid the appearance of 3-dimensional Jacobian matrix, $J$ is reformulated w.r.t. $D_{i,j}$,
  simplified further, given the form of $\text{Cov}[D]$:
\begin{equation}
J^{i,j} = \frac{\partial V_2}{\partial D_{i, j}}(D) \quad J = \xi \sum_{i,j}^{k, 3} J^{i,j}{J^{i,j}}^T
\label{eq:jac}
\end{equation}
Then the intermediate matrix $\Omega^{i,j}_{V_2} \in \mathbb{R}^{\{3 \times 1\}}$ is defined:
\begin{equation}
  \Omega^{i,j}_{V_2} = \begin{bmatrix}
                        \omega_0 \\
                        \omega_1 \\
                        0
                       \end{bmatrix} \quad
  \omega_l = C_l \begin{bmatrix}
                  [U^T\Delta^{(i,j)} V]_{l, 2} \\
                  [U^T\Delta^{(i,j)} V]_{2, l}
                 \end{bmatrix}
\label{eq:omega}
\end{equation}
where $l \in \{0, 1\}$, $C_l = \frac {1} {S_{l,l}^2 - S_{2,2}^2} \cdot \begin{bmatrix}
                                      S_{l,l} & S_{2,2}
                                       \end{bmatrix}$
and $\Delta^{(i,j)}$ defines a matrix, having zeros everywhere except $(i,j)$-th element.
With Eq.~\ref{eq:omega} the Jacobian w.r.t. to $D_{i,j}$ is obtained:
\begin{equation}
  J^{i,j} = -V \Omega^{i,j}_{V_2}, \quad
   \quad 
\end{equation}
For the full derivation of the SVD Jacobian please refer to~\cite{candes2013unbiased},
  ~\cite{papadopoulo2000estimating}.

Recovered in Eq.~\ref{eq:cov_v2}, $\text{Cov}[V_2]$ lies in the sensor coordinate frame,
  hence it should be rotated into an alignment with $V_2$ before the threshold may be applied:
\begin{equation}
  \text{Cov}[V_2] = Q C Q^{-1}
\label{eq:unrotate}
\end{equation}
The matrix $C$ is a covariance of $V_2$ in a normal-aligned coordinate frame
  computed in Eq.~\ref{eq:unrotate} through eigendecomposition.
Having $C$ for an each point in a range scan $\widetilde{X}_i$, the points associated with
  the higher normal uncertainty are discarded: $C_{2, 2} \ge c_\tau$.
The filtration result is a range scan $X_i$ with precise normals, utilized on later stages of the ICP framework.

\subsection{Geometric Correspondence Rejector} \label{sec:geom_rej}
For a current range scan $X_i$ the set of correspondences $\widetilde{M}$ is found
  by a search in the previous range scan $X_{i-1}$
    (Alg.~\ref{alg:icp}, line~\ref{alg:icp:match}).
The proposed model yields filtered correspondences $M = \{ \forall \widetilde{m}_k \in \widetilde{M} : \text{isInlier}(p_k, p'_k, \tilde{m}_k) \}$,
  where $p_k \in X_i$ is a point in the current range scan, participating in the correspondence $\widetilde{m}_k$ with a point $p'_k \in X_{i-1}$ from the previous range scan.

\begin{figure}[h!]
\centering{\includegraphics[width=.9\linewidth]{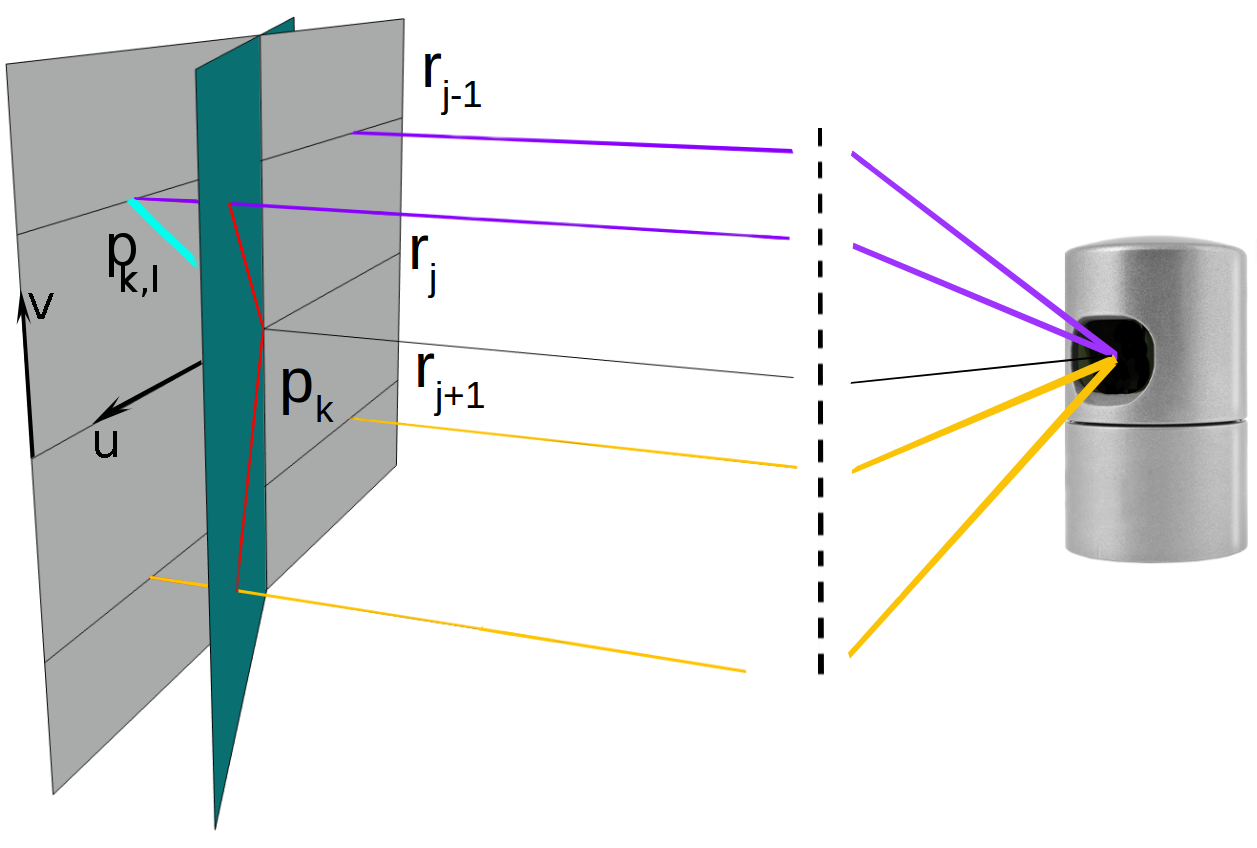}}
  \caption{A laser beam, producing a point $p_k$ (black), it's neighbor beams
    from the upper ring (purple) and the lower ring (orange) together define a plane (gray),
      intersected by the reflecting surface (green), orthogonal to normal at $p_k$.
      The highest \textit{neighbor beam distance}, which is a distance between a point $p_k$ and a point where a neighbor laser beam hits the reflecting surface,
      is used as the rejection threshold.
  }
\label{gr:rejector}
\end{figure}

The test \textit{isInlier} is conducted independently for an every correspondence $\widetilde{m}_k$,
  characterized by the Euclidian distance between matched points: $|| \widetilde{m}_k || = || p_k - p'_k ||$.  
The test description follows:
\begin{equation}
  || \widetilde{m_k} || < \max_{l \in \{0 \dots 3\}}d(p_k, p_l)
\label{eq:rej_thresold}
\end{equation}
The matching distance $|| \widetilde{m}_k ||$ in compared to an upper bound of
  \textit{neighbor beam distances}, $d(p_k, p_l)$
  or, shortly, $d_{k,l}$, highlighted red in Fig.~\ref{gr:rejector}.

The neighbors of $p_k$, $p_l$, $l\in\{0\dots3\}$ are four hypothetical points, which might have been registered by lasers, vertically adjacent to the one registering $p_k$.
Conventionally for spinning lidars, these laser levels are called rings, so if $p_k$ is a part of the ring $r_j$, then it's neighbors $p_l$ are
on the adjacent rings $r_{j-1}$, $r_{j+1}$, shown in purple and orange in Fig.~\ref{gr:rejector}.

The model introduces two neighbors on an each of the adjacent rings assuming that "left" neighbors were registered at a previous increment of lidar azimutal rotation, $p_k$ at a current, and "right" neighbors at a next one.

For the calibrated lidar, all angular distances between $p_k$ and it's neighbors $p_l$ are easily found.
The calibration parameters are:
  $\phi$~(an angular increment of lidar azimutal rotation),
  $\theta_{j-1, j}$, $\theta_{j, j+1}$ are angular distances in pitch between rings.

Hence, $p_k$ with it's neighbors $p_l$ define a segment of the sphere.
This segment could be approximated by a gray plane on Fig.~\ref{gr:rejector},
  in turn assuming neighbor lidar beams parallel.
This assumption was found admissable for lidars in our use, for the hardware parameters are close to $0$: $\phi \approx 0.08^{\circ}$,
  $\theta \approx 0.26^{\circ}$.

The diagonal vector $p_{k,l}$ between $p_k$ and $p_l$ on the approximated plane
  (shown in cyan in Fig.~\ref{gr:rejector}) is obtained as follows:
\begin{equation}
\begin{split}
  & p_k^{\text{ground}} = [{p_k}_x\ {p_k}_y \ 0]^T \ \
    u = p_k \times p_k^{\text{ground}} \ \ 
    v = p_k \times u\\
  & p_{k,l} = [\phi \cdot  u \pm \theta_{j\pm1, j} \cdot v] \cdot || p_k ||
\end{split}
\end{equation}
The vectors $u$, $v$ define absciss and ordinate axes of $p_{k,l}$ coordinate frame.
With $p_{k,l}$ proceed to the \textit{neighbor beam distance}:
\begin{equation}
  d_{k,l} = \frac { || p_{k,l} ||^2 } {\sqrt{|| p_{k,l} ||^2 - ||\langle p_{k,l}, n_p\rangle ||^2 }}
\end{equation}
where $d_{k,l}$ is a projection of $p_{k,l}$ onto the reflecting surface orthogonal to $n_k$ (green surface in Fig.~\ref{gr:rejector}) along the laser beam flight direction.
Matches having $|| \widetilde{m_k} ||$ higher than that are probably made with points laying even further away from $p_k$ than $p_l$ and should be discarded.

\section{Experiments} \label{sec:experiments}
The thorough evaluation of the proposed method is provided.
Two parts of this exist: the dedicated study of the geometric correspondence rejector and 
the integrated test of the proposed lidar odometry.
The former is a comparison of the proposed rejector with the Euclidian distance rejector while
the latter is a test with three ICP odometry variants and the competitor apporach, LOAM, on
two real-world AV benchmarks, renown KITTI and the internal dataset, taken during Yandex own AV fleet operation.

\begin{figure}[h!]
\centering{
  \includegraphics[width=.8\linewidth]{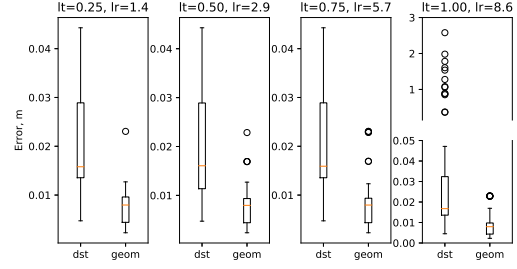}
}
\caption{Translation error of ICP point cloud alignment with different outlier rejectors.
         The proposed in Sec.~\ref{sec:geom_rej} GCR scheme~(geom) outperforms the baseline --
         distance-based rejection~(dst) on every noise level.
         Noise parameters are in meters and degrees. }
\label{gr:rejectors_comparison}
\end{figure}
\subsection{Outlier Correspondence Rejection}
The geometric correspondence rejector~(GCR) is evaluated against the baseline, which is the Euclidian distance rejector.
The test is run on $100$ point cloud pairs from the internal dataset.
The ground truth for the test is relative transformations between pairs, refined by the graph optimization.

In an every test instance the initial guess is the composition of the ground truth transformation with a random noise.
The method's ability to recover the correct transformation is evaluated for several noise levels.
The proposed method have shown the performance, superior to the baseline, as in Fig.~\ref{gr:rejectors_comparison}.
Only distributions of translation estimation errors are included as the rotation errors were found to be very correlated with it.

The noise transformation have two parameters $l_r$, $l_t$~(magnitude limits in rotation and translation) and is sampled as follows:
  generate two random unit vectors on a sphere, where the former is the rotation axis $v_r$ and the latter is the translation axis $v_l$,
  rotate for $m_r \sim U[-l_r \dots l_r]$ about $v_r$ and translate for $m_t \sim U[-l_t\dots l_t]$ along $v_t$.

\subsection{Lidar Odometry Evaluation}
The study of the fully integrated odometry method follows with the baseline~(BL),
  where the Euclidian distance rejector is used within the ICP framework, described in Alg.~\ref{alg:icp}
  along with the common approach for points filtration, following Sec.~\ref{sec:model}.

The baseline method is used as a control in a performance evaluation of
  LOAM, the proposed method~(SALO) and an intermediate variant, integrating NCF into BL.
GCR was not evaluated separately from NCF, because it relies on the quality of point cloud normals,
  which is ensured by NCF.

The internal Yandex dataset of total length of 16 minutes, 12 km was compiled for this study.
It was taken in different cities and seasons of the year with a 32-beam lidar,
mid-grade single-band GNSS with RTK.
GNSS signal frequency was increased by the Kalman filter interpolation with IMU and wheel odometry to be
  used as the ground truth in this evaluation.
The removal of dynamic objects from lidar scans was not performed.

\begin{figure*}[t!]
\centering{\scriptsize {
  \begin{tabular}{cccc}
    \includegraphics[width=.24\linewidth]{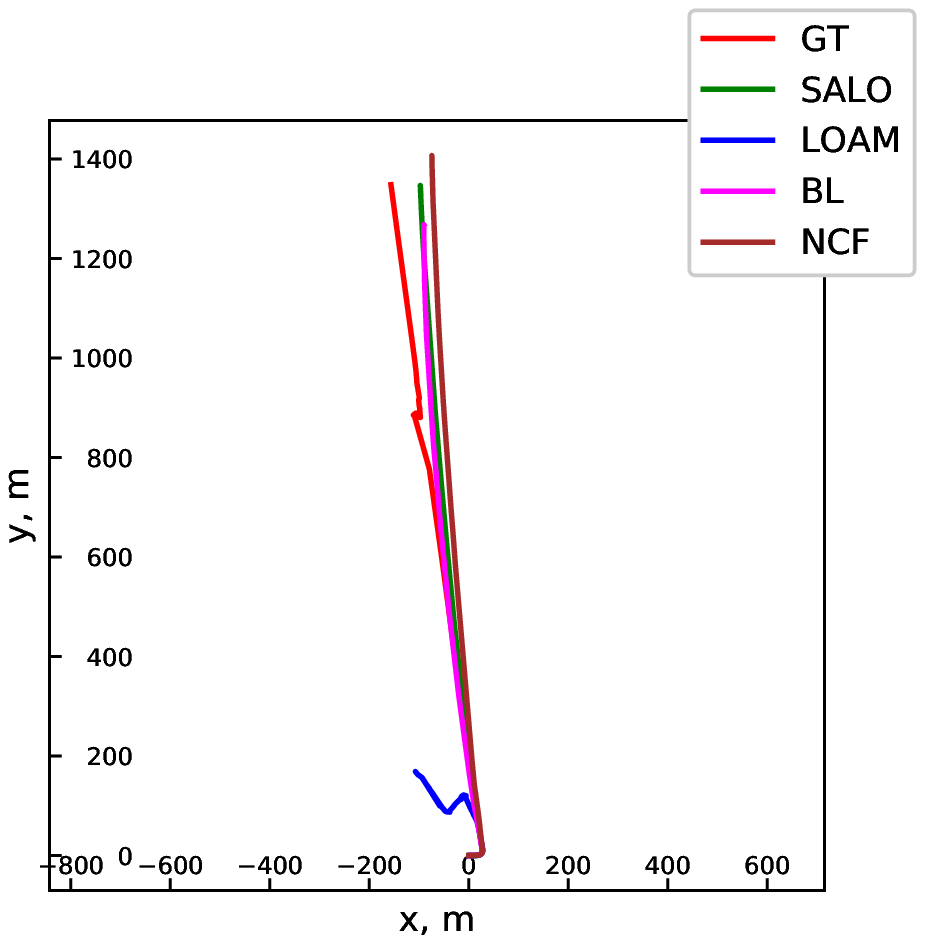} &
    \includegraphics[width=.24\linewidth]{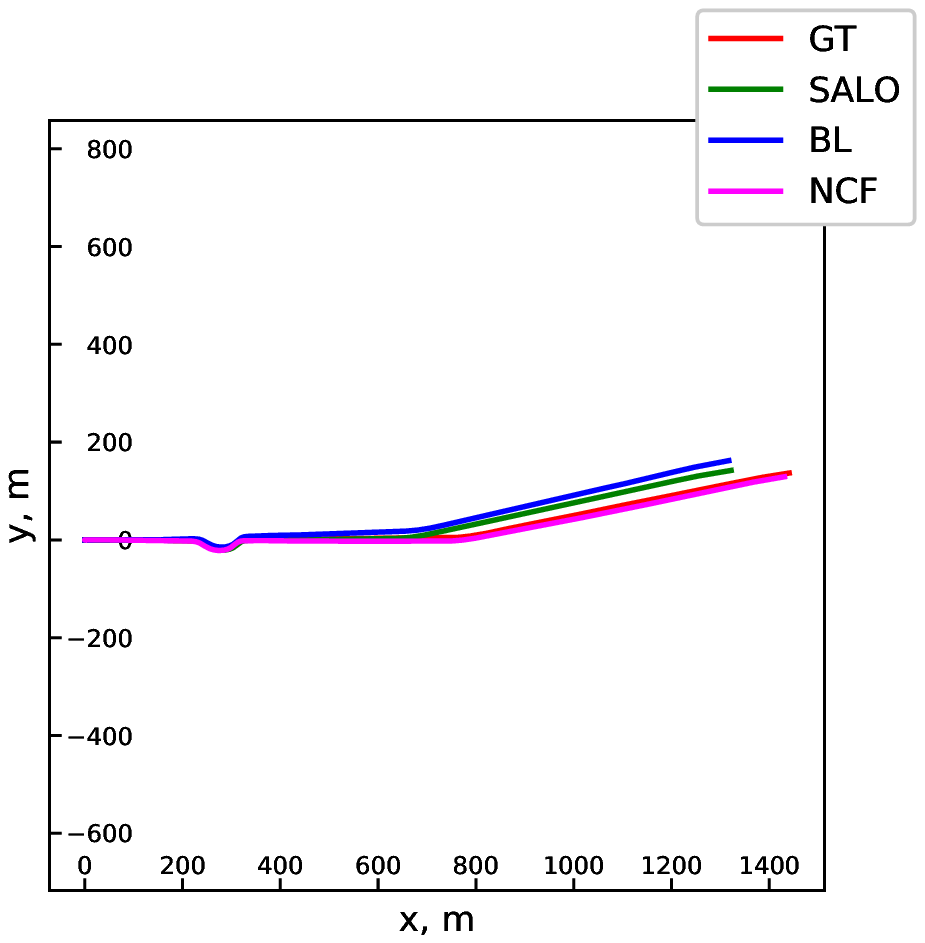} &
    \includegraphics[width=.24\linewidth]{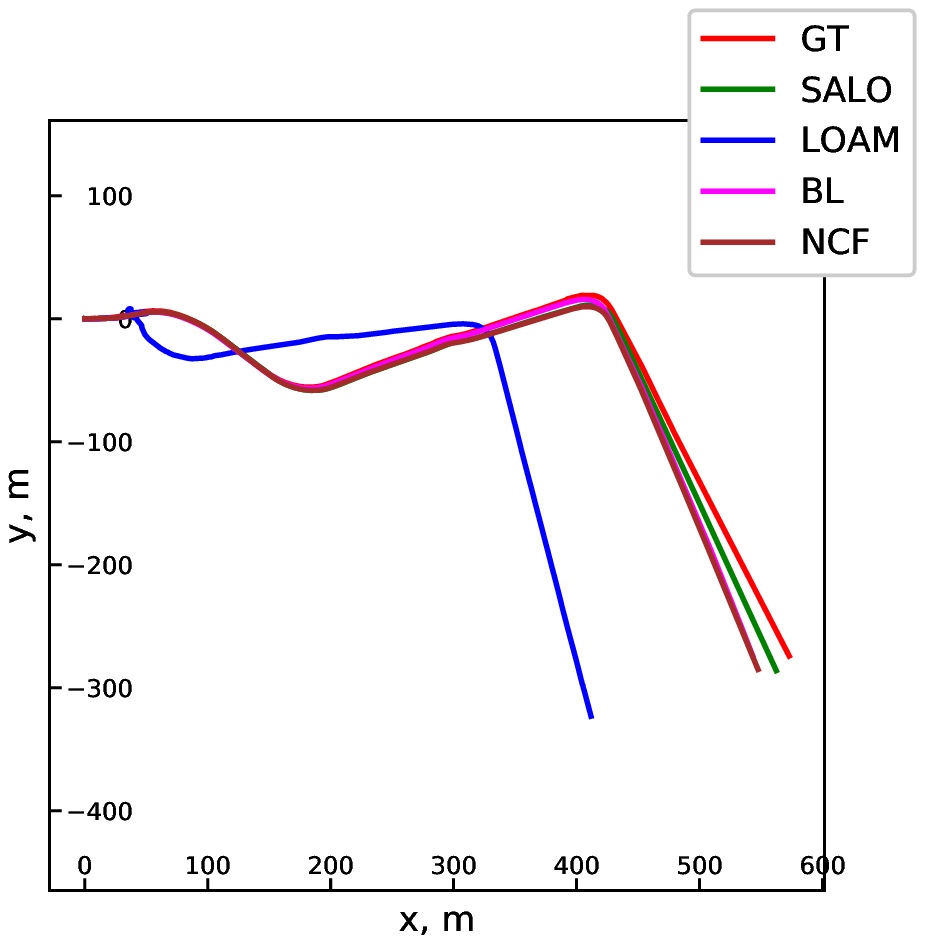} &
    \includegraphics[width=.24\linewidth]{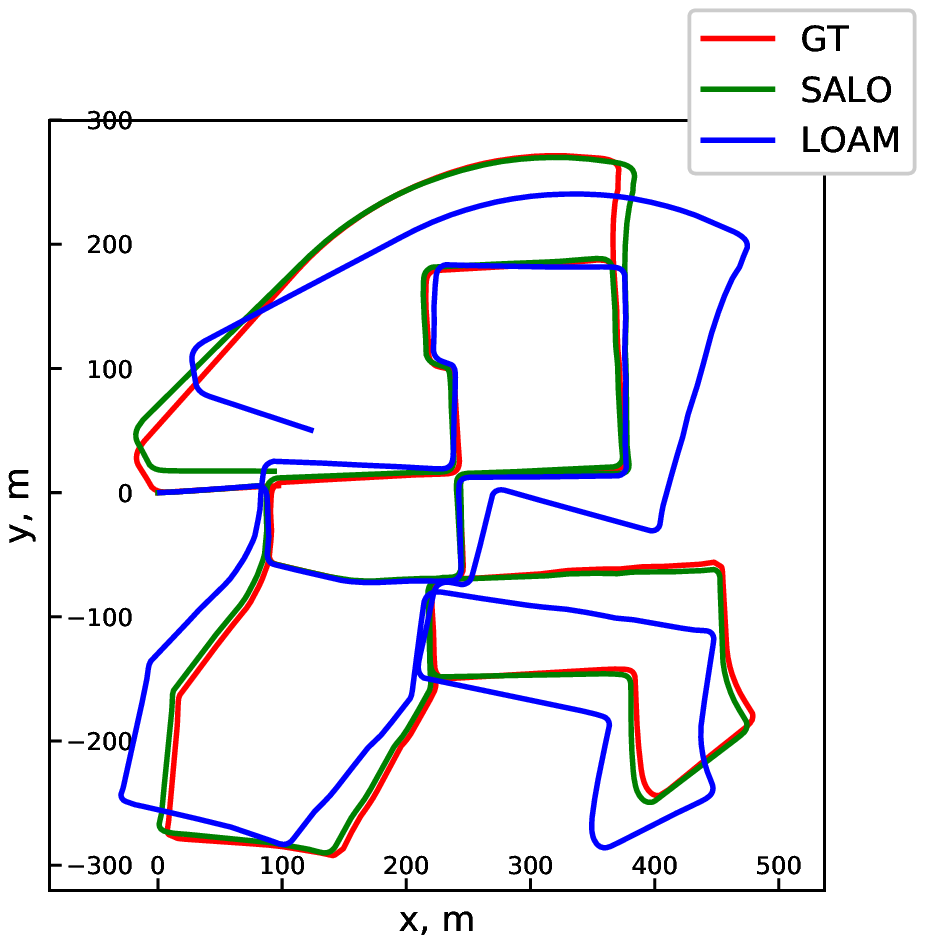} \\
    a) internal \#02 & b) internal \#06 & c) internal \#07 & d) KITTI \#00 \\
  \end{tabular}
  \caption{SALO~(green) is compared to the ground truth~(red), baselines and LOAM on the internal dataset~(a-c) and KITTI~(d).
           SALO shows lower drift in all scenarios except b),
           while the community-supported implementation of LOAM\cite{community_loam_velodyne}
           does not perform on a level reported in literature.
}
\label{gr:sdcdata_trajectories}
}
}
\end{figure*}

Detailed results for all three methods and the competitor are showcased in
  Tab.~\ref{tab:drift} with trajectory visualizations in Fig.~\ref{gr:sdcdata_trajectories}.

A. Geiger argues in~\cite{geiger2012we} that the absolute metric accounts errors non-uniformly
  based on a time passed since the start of the sequence.
We adhere to KITTI's standard and use the same metric and the official devkit in all evaluations,
 with a minor change: the reference pose is set $100$ metres away instead
   of averaging across $50$, $100$ \dots $800$ to foster the results interpretability.
The other motivation for that is space constraints and we may suggest readers
  to refer to our submission on the benchmark website~\cite{geiger2012we} for more results.

Along with error averages, standard deviations are listed and
  it may be noted that the lower mean error always coincides with the lower deviation.

It was shown before~\cite{sun2018dlo} and is confirmed now on the internal dataset and KITTI~(Fig.~\ref{gr:sdcdata_trajectories}.d) that
  the community-supported implementation of LOAM~\cite{community_loam_velodyne} does not perform accordingly to the version,
  competing in the KITTI's leaderboard and tends to degrade in many cases.
Contrary to that, the proposed method produces a trajectory close to the ground truth,
  accumulating little drift over 7 minutes~(Fig.~\ref{gr:sdcdata_trajectories}.d).

The proposed method achieves 27\% and 45\% drift reduction on KITTI and internal dataset respectively
  along with an even stronger error variance reduction.

The avarage processing time of a point cloud is $0.6$~s.
Studied variants do not vary substantially in computational cost,
  which mostly comes from the normal estimation.

\begin{table*}[t!]
\caption{Mean Relative Positioning Error per 100 m}
\centering{ 
\tiny {
\sisetup{detect-weight=true,detect-inline-weight=math}
\setlength{\tabcolsep}{0.3em}
  \begin{tabular}{l|SS|SS|SS|SS|SS|SS|SS|SS|SS|SS|SS|SS}
    \toprule
    \multicolumn{25}{c}{\scriptsize {Training set of KITTI}} \\
    \midrule
   \multirow{2}{*}{Method} &
    \multicolumn{2}{c}{00} &
    \multicolumn{2}{c}{01} &
    \multicolumn{2}{c}{02} &
    \multicolumn{2}{c}{03} &
    \multicolumn{2}{c}{04} &
    \multicolumn{2}{c}{05} &
    \multicolumn{2}{c}{06} &
    \multicolumn{2}{c}{07} &
    \multicolumn{2}{c}{08} &
    \multicolumn{2}{c}{09} &
    \multicolumn{2}{c}{10} &
    \multicolumn{2}{c}{Overall} \\ 
    &
    $\mu\text{, m}$ & $\sigma\text{, m}$ &
    $\mu\text{, m}$ & $\sigma\text{, m}$ &
    $\mu\text{, m}$ & $\sigma\text{, m}$ &
    $\mu\text{, m}$ & $\sigma\text{, m}$ &
    $\mu\text{, m}$ & $\sigma\text{, m}$ &
    $\mu\text{, m}$ & $\sigma\text{, m}$ &
    $\mu\text{, m}$ & $\sigma\text{, m}$ &
    $\mu\text{, m}$ & $\sigma\text{, m}$ &
    $\mu\text{, m}$ & $\sigma\text{, m}$ &
    $\mu\text{, m}$ & $\sigma\text{, m}$ &
    $\mu\text{, m}$ & $\sigma\text{, m}$ &
    $\mu\text{, m}$ & $\sigma\text{, m}$ \\ \midrule
    LOAM &
    2.71  & 2.39 &
    32.64 & 40.19 &
    5.83  & 14.16 &
    2.30  & 1.33 &
    1.12  & 0.70 &
    2.02  & 1.64 &
    2.53  & 2.63 &
    2.75  & 2.13 &
    3.05  & 2.47 &
    2.18  & 1.46 &
    2.43  & 1.90 &
    4.68  & 12.78 \\ 
    BL &
    1.12 & 0.73 &
    4.73 & 8.15 &
    1.18 & 0.53 &
    \bfseries 1.54 & \bfseries 0.44 &
    0.79 & 0.17 &
    0.82 & 0.34 &
    0.60 & 0.16 &
    1.06 & 0.47 &
    1.49 & 1.44 &
    0.92 & 0.42 &
    1.20 & 0.43 &
    1.31 & 2.10 \\ 
    NCF & 
    0.95 & 0.71 &
    2.39 & 6.28 &
    0.99 & 0.46 &
    1.64 & 0.45 &
    0.58 & 0.17 &
    0.59 & 0.40 &
    0.51 & 0.14 &
    \bfseries 0.76 & \bfseries 0.36 &
    \bfseries 1.32 & \bfseries 1.43 &
    0.67 & 0.25 &
    \bfseries 0.95 & \bfseries 0.38 &
    1.02 & 1.61 \\
    SALO &
    \bfseries 0.91 & \bfseries 0.72 &
    \bfseries 1.13 & \bfseries 0.37 &
    \bfseries 0.98 & \bfseries 0.45 &
    1.76 & 0.50 &
    \bfseries 0.51 & \bfseries 0.17 &
    \bfseries 0.56 & \bfseries 0.29 &
    \bfseries 0.48 & \bfseries 0.13 &
    0.83 & 0.51 &
    1.33 & 1.43 &
    \bfseries 0.64 & \bfseries 0.30 &
    0.97 & 0.41 &
    \bfseries 0.95 & \bfseries 0.80 \\ 
    \bottomrule
\end{tabular} \\
}

\vspace{0.45em}

\sisetup{detect-weight=true,detect-inline-weight=math}
\setlength{\tabcolsep}{0.3em}
\begin{tabular}{l|SS|SS|SS|SS|SS|SS|SS|SS|SS|SS}
    \multicolumn{21}{c}{\scriptsize{Internal Dataset}} \\
    \midrule
   \multirow{2}{*}{Method} &
    \multicolumn{2}{c}{00} &
    \multicolumn{2}{c}{01} &
    \multicolumn{2}{c}{02} &
    \multicolumn{2}{c}{03} &
    \multicolumn{2}{c}{04} &
    \multicolumn{2}{c}{05} &
    \multicolumn{2}{c}{06} &
    \multicolumn{2}{c}{07} &
    \multicolumn{2}{c}{08} &
    \multicolumn{2}{c}{Overall} \\ 
    &
    $\mu\text{, m}$ & $\sigma\text{, m}$ &
    $\mu\text{, m}$ & $\sigma\text{, m}$ &
    $\mu\text{, m}$ & $\sigma\text{, m}$ &
    $\mu\text{, m}$ & $\sigma\text{, m}$ &
    $\mu\text{, m}$ & $\sigma\text{, m}$ &
    $\mu\text{, m}$ & $\sigma\text{, m}$ &
    $\mu\text{, m}$ & $\sigma\text{, m}$ &
    $\mu\text{, m}$ & $\sigma\text{, m}$ &
    $\mu\text{, m}$ & $\sigma\text{, m}$ &
    $\mu\text{, m}$ & $\sigma\text{, m}$ \\ \midrule
    LOAM &
    27.57  &  42.64 &
    508.94  &  362.09 &
    63.91  &  38.68 &
    \bfseries 1.96  &  \bfseries 1.24 &
    4.66  &  3.05 &
    145.69  &  116.49 &
    368.47  &  170.30 &
    9.85  &  20.69 &
    7.31  &  7.31 &
    181.22 & 276.75 \\
    BL &
    2.06 & 1.14 &
    2.69 & 1.43 &
    11.32 & 9.88 &
    3.61 & 1.68 &
    50.88 & 64.14 &
    5.97 & 6.66 &
    10.61 & 14.11 &
    5.01 & 3.49 &
    4.63 & 3.57 &
    7.14 & 16.62 \\
    NCF &
    \bfseries 2.02 & \bfseries 1.14 &
    \bfseries 2.72 & \bfseries 1.42 &
    13.75 & 15.17 &
    2.87 & 1.58 &
    25.47 & 32.64 &
    5.91 & 6.66 &
    \bfseries 1.76 & \bfseries 0.74 &
    \bfseries 3.83 & \bfseries 3.93 &
    \bfseries 4.26 & \bfseries 3.70 &
    5.25 & 10.08 \\
    SALO &
    2.06 & 1.14 &
    2.89 & 1.50 &
    \bfseries 6.11 & \bfseries 8.09 &
    3.08 & 1.34 &
    \bfseries 3.63 & \bfseries 2.71 &
    \bfseries 5.82 & \bfseries 6.66 &
    1.82 & 0.73 &
    3.88 & 4.00 &
    4.34 &  3.70 &
    \bfseries 3.67 & \bfseries 4.28 \\ \bottomrule
\end{tabular} \\
\label{tab:drift}
\vspace{0.3em}
\scriptsize{BL is a baseline odometry, as described in Alg.~\ref{alg:icp},
      NCF -- the baseline enhanced with the normal covariance filtration, as in Sec.~\ref{sec:cov_filtr},
      SALO is the proposed method, i.e. BL+NCF+GCR}.
}
\end{table*}

\section{Conclusions} \label{sec:conclusions}
The MIT/Cornell collision during DARPA Grand Urban Challenge is the first AV road incident,
  but regretfully would not be the last.
We share our experience in designing odometry solution for self-driving car
  hoping to foster safety and reliabilty of the autonomous fleets across the globe.
In this work ICP lidar odometry performance boosted by the normal covariance point cloud filter
and by the geometric outlier correspondence rejector.
These routines jointly enable for cleaner point clouds with better correspondences
  yielding better convergence properties, which is demonstrated on KITTI and internal dataset with
  relative positioning errors of $1.37\%$ and $3.67\%$ respectively.
Further pursuit is an implementation of a local map and fast map query mechanisms
  to allow for implicit loop closures.
Another conceived improvement is mixing point-to-point and point-to-plane losses depending on the local structure,
  as well as the preparation of internal dataset for public release.

\thanks{We are grateful to our colleague Sergey Krutikov for his invaluable contribution into this project.}

\printbibliography
\end{document}